\documentclass[11pt]{article} 
\usepackage{rldmsubmit,palatino}
\usepackage{graphicx}
\usepackage{algorithm, algorithmic, caption}
\usepackage{amsmath}
\usepackage{multirow}
\usepackage{wrapfig}

\title{Hierarchical Reinforcement Learning Framework for Adaptive Walking Control Using General Value Functions of Lower-Limb Sensor Signals}

\author{
Sonny T. ~Jones \\
Department of Biomedical Engineering\\
University of Utah\\
\texttt{sonny.jones@utah.edu} \\
\And
Grange M. Simpson \\
Department of Biomedical Engineering\\
University of Utah\\
\texttt{grange.simpson@utah.edu} \\
\And
\hspace{22pt} Patrick M. Pilarski \\
\hspace{22pt}Alberta Machine Intelligence Institute (Amii)\\
\hspace{22pt}Department of Medicine\\
\hspace{22pt}Department of Computing Science\\
\hspace{22pt}University of Alberta\\
\hspace{22pt}\texttt{pilarski@ualberta.ca} \\
\And
\hspace{-13pt} Ashley N. Dalrymple \\
\hspace{-13pt} Department of Biomedical Engineering\\
\hspace{-13pt} Department of Physical Medicine and Rehabilitation\\
\hspace{-13pt} University of Utah\\
\hspace{-13pt} \texttt{ashley.dalrymple@utah.edu} \\
}

\begin{document}

\maketitle

\begin{abstract}
Rehabilitation technology is a natural setting to study the shared learning and decision-making of human and machine agents. In this work, we explore the use of Hierarchical Reinforcement Learning (HRL) to develop adaptive control strategies for lower-limb exoskeletons, aiming to enhance mobility and autonomy for individuals with motor impairments. Inspired by prominent models of biological sensorimotor processing, our investigated HRL approach breaks down the complex task of exoskeleton control adaptation into a higher-level framework for terrain strategy adaptation and a lower-level framework for providing predictive information; this latter element is implemented via the continual learning of general value functions (GVFs). GVFs generated temporal abstractions of future signal values from multiple wearable lower-limb sensors, including electromyography, pressure insoles, and goniometers. We investigated two methods for incorporating actual and predicted sensor signals into a policy network with the intent to improve the decision-making capacity of the control system of a lower-limb exoskeleton during ambulation across varied terrains. As a key result, we found that the addition of predictions made from GVFs increased overall network accuracy. Terrain-specific performance increases were seen while walking on even ground, uneven ground, up and down ramps, and turns, terrains that are often misclassified without predictive information. This suggests that predictive information can aid decision-making during uncertainty, e.g., on terrains that have a high chance of being misclassified. This work, therefore, contributes new insights into the nuances of HRL and the future development of exoskeletons to facilitate safe transitioning and traversing across different walking environments.
\end{abstract}

\keywords{
Hierarchical Reinforcement Learning, Continual Deep Learning, General Value Functions, Exoskeletons, Walking
}

\acknowledgements{This work was supported by the Departments of Biomedical Engineering and Physical Medicine and Rehabilitation. STJ was supported by the Campbell Endowed Graduate Fellowship from the College of Engineering. PMP was supported by NSERC, Amii, and the Canada CIFAR AI Chairs program.}  

\startmain 

\section{Introduction}

Lower-limb exoskeletons have the potential to aid in the rehabilitation of walking for individuals who have suffered a variety of neuromuscular deficits, including stroke and spinal cord injury \cite{luo2023robust}. Day-to-day community ambulation requires the ability to modify walking to adapt to the characteristics of the environment and walking terrain \cite{balasubramanian2014walking}. However, current exoskeleton controllers are not adaptive, relying on manual physical input by the user to switch to different fixed control strategies, limiting their effectiveness in real-world use \cite{baudexoreview}. There is a need to develop adaptive exoskeleton controllers to aid in safe transitioning and traversing across variable terrains. The development of these control strategies for assistive devices can enhance the overall autonomy and mobility of individuals who suffer from motor impairments.

Hierarchical Reinforcement Learning (HRL) offers a promising approach to simplifying complex tasks by breaking them down into multiple simpler sub-tasks that are adaptable over time. Previous work has utilized an architecture consisting of general value functions (GVFs) and RL agents for HRL to make gesture control decisions in an Android environment \cite{comanici2022learning}. We employ an HRL approach to reduce the control problem into two components: 1) a higher-level framework to adaptively predict which terrain an individual is walking on, and 2) a lower-level framework that provides predictive information of lower-limb sensor signals in the form of GVFs to the higher-level framework to make control decisions. GVFs produce temporal abstractions of future signal values of multiple lower-limb sensors during walking, including electromyography (EMG), underfoot pressure, and knee joint angles. 

Prior work has shown that predictions from GVFs can act as inputs for a policy network \cite{DBLP:conf/icorr/PilarskiDS13} that controls the actions of a learning agent. However, integrating these predictions into the latent layers of a deep neural network might integrate abstract predicted information more effectively and improve decision-making \cite{DBLP:conf/aaai/SherstanDM0P20}. We examine two hierarchical methods for integrating actual and predicted signals into a policy network. The first method involves inputting actual and predicted sensor signals together in the front of the policy network. The second method adds predictions into the latent space after encoding actual signals into the network input. We aimed to determine which approach leads to better performance in predicting the terrain an individual is walking across. This can create adaptive control policies for walking over different terrains using assistive devices.

\section{Methods}

Ten (24.7 $\pm$ 3.2 years old, 3 female) individuals wore a suite of sensors while walking over different terrains. Sensors included EMG for muscle activity, goniometers for knee joint angles, and pressure-sensing insoles for weight distribution under the foot. EMG data were filtered with a 4th-order Butterworth bandpass filter (high pass cut-off at 10 Hz, low pass cut-off at 450 Hz). The goniometer signals were filtered with a 2nd-order low-pass filter at 5 Hz. Signals were then down-sampled to 33 Hz and normalized between 0 and 1, resulting in 30 sensor signals to form the state space. Terrains included even ground, uneven ground, up and down stairs, up and down ramps, and turns, with transitions marked synchronously with data collection. Each participant dataset resulted in approximately 14,000--18,000 timesteps, or approximately 425--545 gait cycles.

Data were sequentially input into the learning architecture to simulate online learning. The first task involved learning predictions of walking-related sensor signals approximately 0.5 seconds into the future (17 timesteps at 33 Hz). GVFs were learned on-policy using true online temporal-difference learning (TOTD), with $\gamma$ set to 0.94. The second task involved learning terrain predictions using actual and predicted signals, with an initial learning rate of $\alpha$ = 0.001. We describe the specific computational methods used and the choices in the subsections below.

{\bf Selective Kanerva Coding (SKC)}: Function approximation methods are required to represent a continuous state space, such as a range of sensor signal values. We randomly distributed $K$ = 5000 prototypes in the 30-dimensional state space. The locations of these prototypes were held constant. The 500, 100, and 25 closest prototypes to the current state became "active", or set to one, to indicate the location of the current state (defined by the values of the sensor signals) within the state space \cite{DBLP:conf/icorr/TravnikP17}. Using these sets of closest prototypes to the current state provides coarse and fine representations of the state \cite{dalrymple2020pavlovian}. The prototype activations were outputted as a binary feature vector and used to generate the GVFs.

\begin{figure}[t!]
    \centering
    \includegraphics[scale = 1]{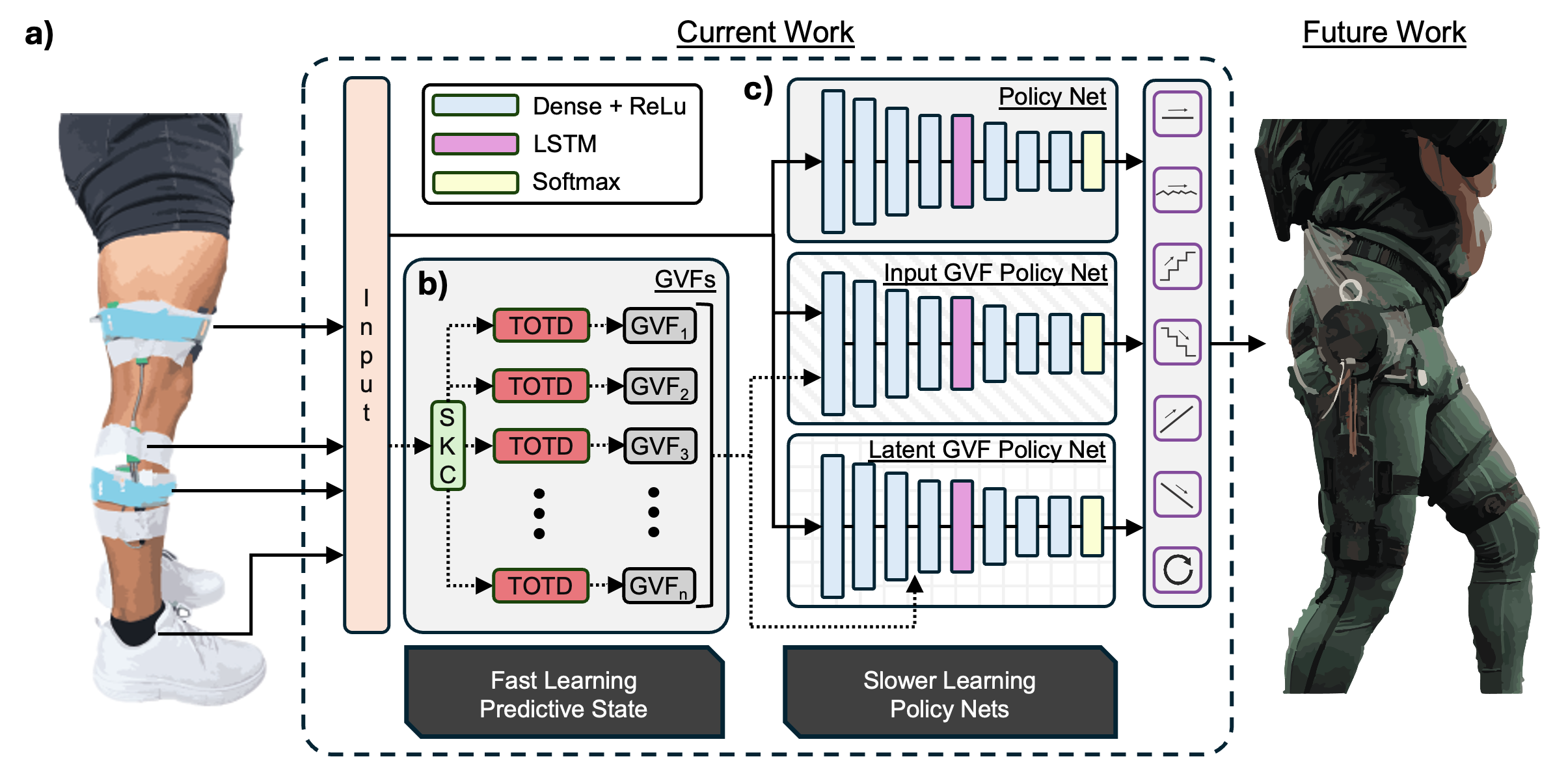}
    \caption{Flowchart of the HRL architecture. a) Overall system diagram showing the flow from lower-limb sensor signal acquisition through prediction algorithms and terrain decision networks to the resulting exoskeleton control policy, b) GVFs used for predicting future sensor signals, c) Three policy network configurations illustrating where GVF predictions are integrated. Exoskeleton image adapted from the EPIC Lab, Georgia Tech.}
    \label{networkarch}
\end{figure}

{\bf General Value Functions (GVFs)}: RL agents use a value function to approximate the expected return, the future cumulative discounted sum of rewards an agent can expect from a particular state \cite{DBLP:books/lib/SuttonB98}. However, value functions can approximate any signal of interest, called a cumulant ($Z$) \cite{white2015thesis, DBLP:conf/atal/SuttonMDDPWP11}. Value functions used in this fashion are called GVFs. A GVF learned online can be used to ask questions regarding the current behavior of robotic systems, making temporally abstracted predictions of the cumulative discounted signal. GVFs utilize a $\gamma$ value between 0 and 1 to represent how far the abstractions are made in the future. These timesteps can be calculated using the equation: $timesteps = \frac{1}{1 - \gamma}$. A GVF with $\gamma$ = 0.94 will predict 17 timesteps into the future. With regard to learning GVFs of sensor signals, the prediction learning task required predicting gait-related signals in real-time. This was done using GVFs, with SKC representing the continuous state space and TOTD (described below) updating the predictions made by the GVFs. This framework allows the GVFs to rapidly learn and predict signals from the lower limb, which we call the "fast predictive state" (Figure \ref{networkarch}b).

{\bf True Online Temporal-difference (TOTD) Learning}: Temporal-difference (TD) learning is a core algorithm in RL, allowing agents to learn from their experiences without needing a model of the environment. \cite{sutton1988learning}. TD learning has been used to learn GVFs due to their low computational cost and performance \cite{DBLP:journals/jmlr/SeijenMPMS16}. TOTD is an up-to-date TD learning algorithm that follows the equivalent ideal mathematical outcomes. It has superior online performance in many settings compared to other TD learning methods \cite{DBLP:journals/jmlr/SeijenMPMS16}. We set $\lambda$, the eligibility trace parameter, to 0.5 based on previously tested values.

{\bf Policy Network for Terrain Control Decisions}: We updated the policy networks during training using actual and predicted gait-related signals to classify the appropriate terrain in real-time. For this task, we employed a continual deep learning network that utilized a replay buffer to sample previously acquired samples as batch inputs to our network \cite{DBLP:conf/aaai/SherstanDM0P20, DBLP:conf/nips/RolnickASLW19}. We adopted the sampling strategy from the CLEAR framework to minimize catastrophic forgetting, which creates new batches for continual deep learning by allocating portions of the batch for new and replayed data \cite{DBLP:conf/nips/RolnickASLW19}.

The policy network was updated with a 50-50 new-replay batch of size 32 samples, where half of the data came from the most recently collected samples, and the other half was randomly sampled from a replay buffer of size 1000. This approach ensured that the network benefited from both recent and past experiences. We deem the policy network as the "slower learning policy net", as deep learning architectures typically require more iterations and time to learn compared to the RL methods applied here.

Actual signals were acquired and input into the fast predictive state to obtain the predictions made by the GVFs. We compared embedding GVFs directly into the input stream compared to embedding GVFs into the network's latent space, and compared both to a control net without GVFs (only actual sensor signals). All policy networks (nets) used a series of decreasing-size encoding layers to transform the original data into compressed latent representations \cite{DBLP:conf/aaai/SherstanDM0P20}. For the Input GVF Policy Net, the GVF predictions were directly appended to the actual signals and both were input into the network. For the Latent GVF Policy Net, the GVF predictions were appended to the post-encoding layer to the output of the actual signal input encoding. The Policy Net with no GVFs as inputs was trained as a control net. The architecture for the policy networks are shown in Figure \ref{networkarch}c.

{\bf Statistics: } Data normality was assessed using the Shapiro-Wilk test. Due to the non-normal distribution, we used the Kruskal-Wallis test to compare accuracies between all three policy nets. We utilized Dunn's test with the Holm-Sidak correction to determine significant differences between individual terrain predictions between networks.

\begin{minipage}{0.49\textwidth}
\vspace{-12pt}
\begin{algorithm}[H]
\small
\captionsetup{labelfont={sc,bf}}
\caption{{\small Selective Kanerva Coding} \\ {\small $K$ = number of prototypes; $n$ = number of sensors; $c$ = closest prototypes to current state; $P$ = prototypes; $S$ = state; $D$ = distance vector; $x$ = feature vector}}
\begin{algorithmic}[t]
\STATE Parameters: $K$, $c$
\STATE Initialize $K$ prototypes in state space, $D \leftarrow zeros(K)$
\STATE Input new state S'
\FOR{$i$ = 1 to $K$}
    \FOR{$j$ = 1 to $n$}
        \STATE $D_i \leftarrow d(P_{i,j}, S_j)$ \hfill \COMMENT{d = Euclidean distance}
    \ENDFOR
\ENDFOR
\STATE $I \leftarrow Quickselect(D)$ \hfill \COMMENT{Indices}
\FOR{$m$ = 1 to 3}
    \STATE $ind_m \leftarrow I(1\,to\,c_m)$
\ENDFOR
\STATE $x_{indm} \leftarrow 1$ \hfill \COMMENT{Offest (m - 1) * K}
\end{algorithmic}
\end{algorithm}
\vspace{-10pt}
\end{minipage}
\hfill
\begin{minipage}{0.49\textwidth}
\vspace{-20pt}
\begin{algorithm}[H]
\small
\captionsetup{labelfont={sc,bf}}
\caption{{\small True Online Temporal Difference Learning} \\ {\small $w$ = weight vector; $e$ = eligibility trace; $V$ = general value function; $S$ = state; $x$ = feature vector; $Z$ = cumulant; $\delta$ = temporal difference; $\gamma$ = termination signal/discounting factor; $\lambda$ = eligibility trace parameter; $\alpha$ = learning rate}}
\begin{algorithmic}[t]
\STATE Input: ${\lambda}, {\gamma}, {\alpha}$
\STATE Initialize: $w, x, V_{old}, S, e \leftarrow 0$
\STATE Repeat every time-step:
\STATE Generate next state $S'$ and cumulant $Z'$
\STATE $x' \leftarrow SKC(S')$
\STATE $V \leftarrow w^{T}x$
\STATE $V' \leftarrow w^{T}x'$
\STATE $\delta \leftarrow Z + {\gamma}V' - V$
\STATE $e \leftarrow {\gamma}{\lambda}e + x - {\alpha}{\gamma}{\lambda}(e^{T}x)x$
\STATE $w \leftarrow w + {\alpha}(\delta \; + \; V - V_{old})e \; - \; {\alpha}(V - V_{old}) x$
\STATE $V_{old} \leftarrow V', x \leftarrow x'$ 
\end{algorithmic}
\end{algorithm}
\vspace{-10pt}
\end{minipage}

\section{Results}

\begin{figure}[t!]
  \begin{center}
    \includegraphics[trim = {0, 0.6cm, 0, 0.3cm}, scale = 1]{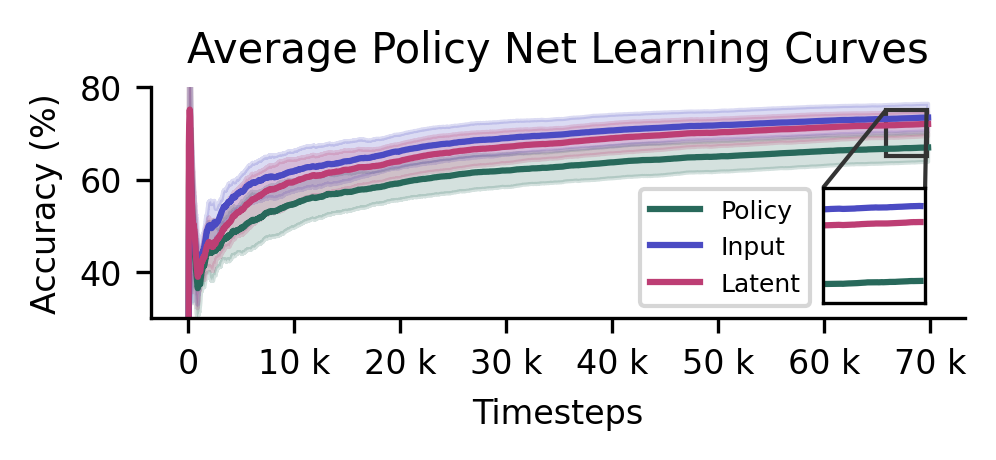}
  \end{center}
  \caption{Convergence curves for policy nets.}
  \label{convergencecurves}
\end{figure}

Figure \ref{convergencecurves} displays the convergence curves of all policy nets. The nets exhibit an initial increase in accuracy when evaluated on the first terrain, since the learner had only been exposed to a single terrain type. As additional terrains were encountered, overall accuracy declined before converging once all terrain varieties were encountered. At the end of training, the control Policy Net achieved a total accuracy across all terrains of 66.9\% ($\pm$ 3.1\%), the Input GVF Policy Net achieved 73.4\% ($\pm$ 3.0\%), and the Latent GVF Policy Net achieved 72.0\% ($\pm$ 2.7\%). Both GVF policy nets achieved significantly higher end-of-training accuracies than the control Policy Net (Input GVF vs. Policy Net: $p = 0.001$, Latent GVF vs. Policy Net: $p = 0.01$). The total accuracies between the two GVF policy nets were not statistically different ($p = 0.42$).

\begin{figure}[t!]
    \centering
    \includegraphics[trim = {0, 0.3cm, 0, 0.3cm}, scale = 0.98]{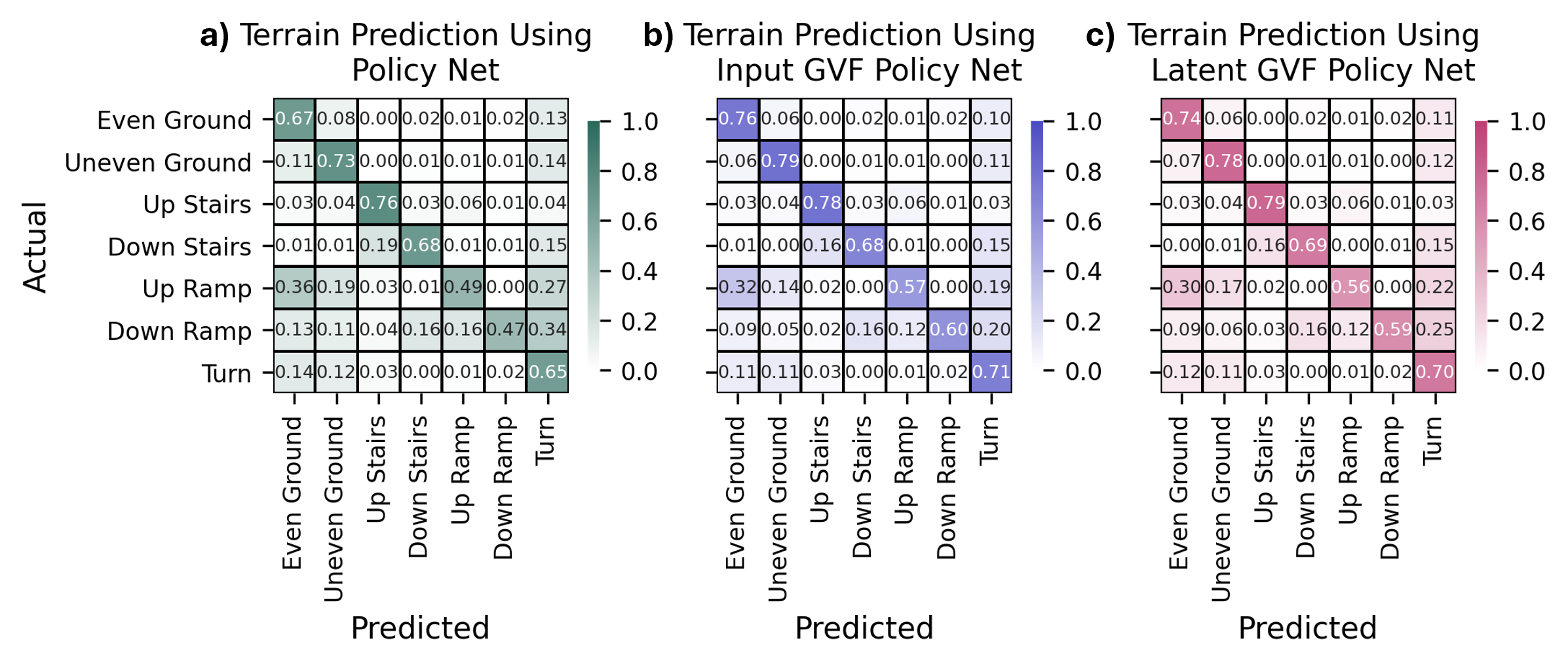}
    \caption{Confusion matrices of the terrain classifications made by the a) Policy Net, b) Input GVF Policy Net, and c) Latent GVF Policy Net, showing correct and predicted labels.}
    \label{confusionmatrix}
\end{figure}

Figure \ref{confusionmatrix} presents the confusion matrices of the terrains classified by all three policy nets. Table 1 reports the average classification accuracies across different terrain types. The Input GVF Policy Net improved classification accuracy for even ground, uneven ground, up ramp, down ramp, and turns by an average increase of 8.3\%, with the largest performance increase of 13.5\% ($p = 0.007$) seen in down ramp classification. The Latent GVF Policy Net also demonstrated similar

\clearpage


improvements across the same terrains, with an average increase of 6.9\% and a peak performance gain of 12.2\% ($p = 0.02$), also seen in down ramp classification. For reference, the baseline control Policy Net achieved a classification accuracy of 46.7\% for down ramp. Both the Input GVF Policy Net and Latent GVF Policy Net showed non-significant improvements in up stairs ($p = 0.58$) and down stairs classification ($p = 0.30$). Overall, these results suggest that integrated GVF Policy Nets enhance terrain classification performance, particularly for challenging terrains such as down ramp. However, not all increases in accuracy were significant, and further analysis is needed to determine the optimal method for integrating GVFs into the network.

\begin{table}[H]
\vspace{-5pt}
\scriptsize
\centering
\caption{Policy Net Average Accuracies Per Terrain}
\vspace{-5pt}
\label{table:1}
\begin{tabular}{ c || c | c | c | c | c | c| c ||} 
 \cline{2-8} 
 & \multicolumn{7}{|c||}{\textbf{Average Accuracies (\%)}} \\ 
 \hline
 \multicolumn{1}{||c||}{\textbf{Network}} & Even Ground & Uneven Ground & Up Stairs & Down Stairs & Up Ramp & Down Ramp & Turns \\ 
 \hline\hline
 \multicolumn{1}{||c||}{Policy Net} & 67.5 ($\pm$ 4.5) & 73.3 ($\pm$ 4.2) & 76.5 ($\pm$ 5.3) & 68.1 ($\pm$ 1.8) & 49.4 ($\pm$ 6.5) & 46.7 ($\pm$ 7.7) & 65.0 ($\pm$ 2.7) \\ 
 \hline
 \multicolumn{1}{||c||}{Input GVF Policy Net} & 75.6 ($\pm$ 3.7)** & 79.2 ($\pm$ 4.1)* & 78.4 ($\pm$ 5.3) & 68.3 ($\pm$ 2.5) & 56.6 ($\pm$ 5.6)* & 60.2 ($\pm$ 7.6)** & 71.5 ($\pm$ 1.9)*** \\ 
 \hline
 \multicolumn{1}{||c||}{Latent GVF Policy Net} & 73.8 ($\pm$ 3.6)* & 78.0 ($\pm$ 3.6) & 78.9 ($\pm$ 5.1) & 69.4 ($\pm$ 2.0) & 55.8 ($\pm$ 5.2)* & 59.0 ($\pm$ 8.3)* & 69.7 ($\pm$ 2.1)** \\ 
 \hline
 \multicolumn{8}{c}{\scriptsize Comparisons made between Policy Net and specific GVF Net. *: $p<0.05$, **: $p<0.01$, ***: $p<0.001$} \\
\end{tabular}
\vspace{-5pt}
\end{table}

\section{Conclusion}

In this work, we implemented hierarchical RL and compared the performance of different policy networks with and without the input of predictions made by GVFs to predict the walking terrain using body-worn sensor signals. Specifically, we investigated whether it is more effective to input these predictions directly into the input layer of the network or the latent space of the post-encoding layers. Preliminary results indicate that adding predictive signal information from GVFs improves classification performance. However, it remains inconclusive whether adding these predictions to the input or latent layers is superior because significant accuracy increases were observed in only some terrains. This implies that predictive information can benefit decision making under uncertainty, for instance, for terrains prone to misclassifications. Future work will utilize GVF-integrated policy nets to create adaptive controllers for exoskeletons.


\end{document}